\documentclass{article}
\usepackage{arxiv}
\usepackage{tikz}
\usepackage{pgfplots}
\usepackage{amsmath,amssymb,amsfonts}
\usepackage{algorithmic}
\usepackage{mwe}
\usepackage{textcomp}
\usepackage{xcolor}
\usepackage{mdframed}
\usepackage{balance}
\usepackage[caption=false, font=footnotesize]{subfig}
\usepackage[export]{adjustbox}
\usepackage{multirow}
\usepackage[nolist]{acronym}
\usepackage[utf8]{inputenc}
\usepackage{hyperref}
\usepackage{cleveref}
\usepackage{bm}
\usetikzlibrary{spy}
\usetikzlibrary{automata,positioning,arrows}
\usepackage{booktabs}
\usepackage{ragged2e}
\usepackage{graphicx}
\usepackage{tikzscale}
\usepackage{natbib}
\setcitestyle{authoryear}
\usepackage{float}

\pgfplotsset{compat=1.5}
\usepgfplotslibrary{groupplots}
\usetikzlibrary{patterns}

\newcommand{\Eg}{E.\,g.}
\newcommand{\eg}{e.\,g.}

\newcommand{\ie}{i.\,e.}

\newcommand{\map}{\mu}
\newcommand{\fmap}{\mathcal{M}}
\newcommand{\amp}{\mathcal{A}}
\newcommand{\ampp}{\mathcal{A}_p}
\newcommand{\Camp}{\mathcal{A}^{C}_{p}}
\newcommand{\pind}{\Phi}

\newcommand{\scampind}[1]{\mathcal{A}^{C}_{p}(#1)'}
\newcommand{\nmap}{\mu_N}

\begin{acronym}
    \acro{fft}[FFT]{Fast Fourier Transform}
    \acro{2fft}[2D FFT]{2D Fast Fourier Transform}
    \acro{dft}[DFT]{Discrete Fourier Transform}
    \acro{idft}[IDFT]{Inverse Discrete Fourier Transform}
    \acro{2dft}[DFT]{Discrete Fourier Transform}
    \acro{gmm}[GMM]{Gaussian Mixture Model}
    \acro{rose}[ROSE]{RObust frequency-based Structure Extraction}
    \acro{em}[EM]{Expectation Maximisation}
\end{acronym}

\title{Robust Frequency-Based Structure Extraction}
\author{Tomasz Piotr Kucner\\
Mobile robotics and Olfaction Lab (MRO)\\
Centre for Applied Autonomous Sensor Systems (AASS)\\
\"{O}rebro University\\
Sweden
\And
Matteo Luperto\\
Applied Intelligent System lab (AISLab)\\
Università degli Studi di Milano Milano\\
Italy
\And
Stephanie Lowry\\
Mobile robotics and Olfaction Lab (MRO)\\
Centre for Applied Autonomous Sensor Systems (AASS)\\
\"{O}rebro University\\
Sweden
\And
Martin Magnusson\\
Mobile robotics and Olfaction Lab (MRO)\\
Centre for Applied Autonomous Sensor Systems (AASS)\\
\"{O}rebro University\\
Sweden
\And
Achim J. Lilienthal\\
Mobile robotics and Olfaction Lab (MRO)\\
Centre for Applied Autonomous Sensor Systems (AASS)\\
\"{O}rebro University\\
Sweden
}

\begin{document}
    \maketitle
    \begin{abstract}
        State of the art mapping algorithms can produce high-quality maps.
        However, they are still vulnerable to clutter and outliers which can affect map quality and in consequence hinder the performance of a robot, and further map processing for semantic understanding of the environment.
        This paper presents ROSE, a method for building-level structure detection in robotic maps.
        ROSE exploits the fact that indoor environments usually contain walls and straight-line elements along a limited set of orientations.
        Therefore metric maps often have a set of {\em dominant directions}.
        ROSE extracts these directions and uses this information to segment the map into {\em structure} and {\em clutter} through filtering the map in the frequency domain (an approach substantially underutilised in the mapping applications).
        Removing the clutter in this way makes wall detection (e.g. using the Hough transform) more robust.
        Our experiments demonstrate that (1) the application of ROSE for decluttering can substantially improve structural feature retrieval (e.g., walls) in cluttered environments, (2) ROSE can successfully distinguish between clutter and structure in the map even with substantial amount of noise and (3) ROSE can numerically assess the amount of structure in the map.
    \end{abstract}

    \section{Introduction}
    \label{sec:introduction}
    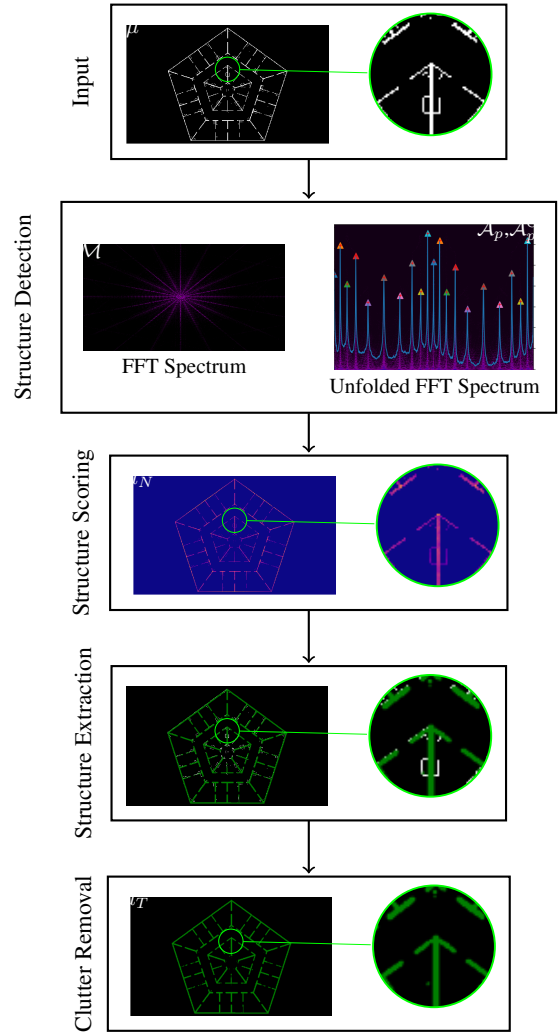
\begin{figure}[H]
        \centering
        \begin{tikzpicture}[shorten >=1pt,on grid,auto,every node/.style={inner sep=2pt,outer sep=0},thick,scale=0.9, every node/.style={scale=0.9}]

            \node[draw,align=center](in){
                \begin{tikzpicture}[spy using outlines={connect spies}]
                    \node{\includegraphics[width=0.2\textwidth]{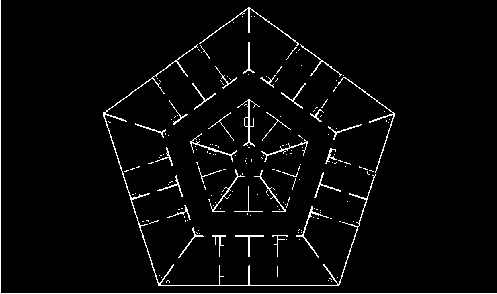}};
                    \spy [circle,green,magnification=5,size=2cm] on (0,0.25) in node at (3,0.15);
                    \node[text=white]at(-1.4,0.8){$\mu$};
                \end{tikzpicture}
            };
            \node[rotate=90, left=3cm of in] {Input};

            \node[draw,below =3cm of in,align=center](sd){
                \begin{tikzpicture}
                    \node(ps){\includegraphics[width=0.2\textwidth,clip,trim=20mm 15mm 15mm 15mm]{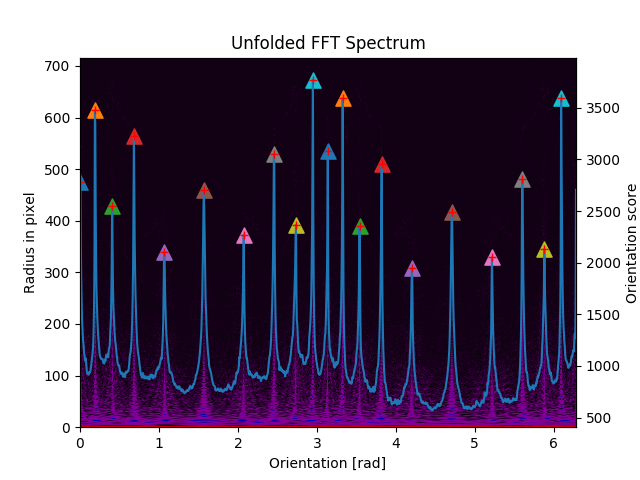}\\
                    Unfolded FFT Spectrum
                    };
                    \node[left=3.7cm of ps]{\includegraphics[width=0.2\textwidth,clip,trim=20mm 15mm 15mm 15mm]{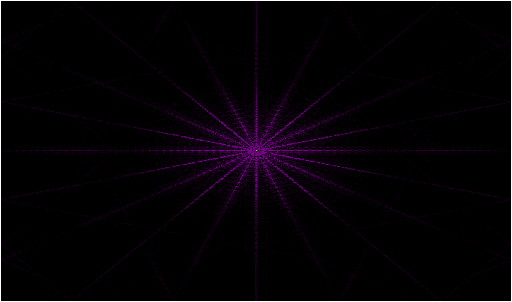}\\
                    FFT Spectrum };
                    \node[text=white]at(-5.1,0.9){$\mathcal{M}$};
                    \node[text=white]at(1.1,1.2){$\mathcal{A}_{p}$,$\mathcal{A}^{C}_{p}$};
                \end{tikzpicture}};
            \node[rotate=90, left=3.8cm of sd] {Structure Detection};

            \node(ss)[draw,below = 3cm of sd,align=center]{
                \begin{tikzpicture}[spy using outlines={connect spies}]
                    \node{\includegraphics[width=0.2\textwidth]{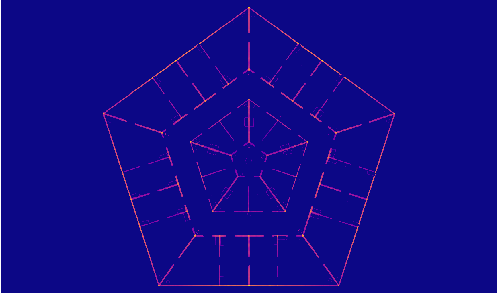}};
                    \spy [circle,green,magnification=5,size=2cm] on (0,0.25) in node at (3,0.15);
                    \node[text=white]at(-1.4,0.8){$\mu_N$};
                \end{tikzpicture}};
            \node[rotate=90, left=3cm of ss] {Structure Scoring};

            \node(se)[draw,below = 2.8cm of ss,align=center]{\begin{tikzpicture}[spy using outlines={connect spies}]
                                                                 \node{\includegraphics[width=0.2\textwidth]{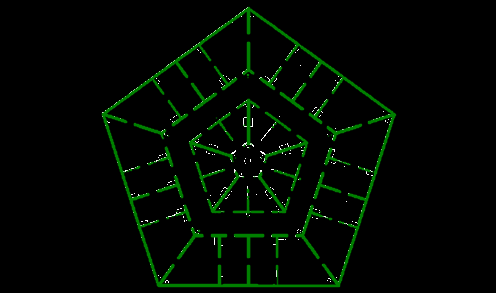}};
                                                                 \spy [circle,green,magnification=5,size=2cm] on (0,0.25) in node at (3,0.15);

            \end{tikzpicture}
            };
            \node[rotate=90, left=3cm of se] {Structure Extraction };
            \node(cr)[draw,below = 2.8cm of se,align=center] {\begin{tikzpicture}[spy using outlines={connect spies}]
                                                                  \node{\includegraphics[width=0.2\textwidth]{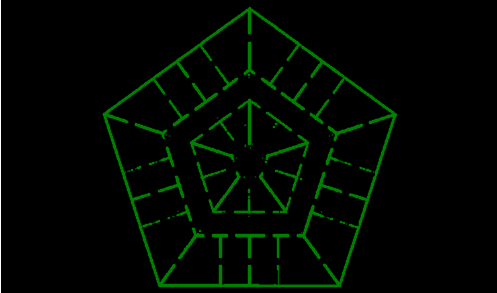}};
                                                                  \spy [circle,green,magnification=5,size=2cm] on (0,0.25) in node at (3,0.15);
                                                                  \node[text=white]at(-1.4,0.8){$\mu_T$};
            \end{tikzpicture}
            };
            \node[rotate=90, left=3cm of cr] {Clutter Removal};
            \draw[->] (in) edge [node distance=6cm] (sd);
            \draw[->] (sd) edge (ss);
            \draw[->] (ss) edge (se);
            \draw[->,] (se) edge (cr);
        \end{tikzpicture}
        \caption{Example of structure scoring applied for clutter removal.
            The \emph{input} map (top) contains structure elements as well as clutter.
            Our structure \emph{detection} method identifies the 10 dominant directions in the map from the 20 symmetric peaks in the frequency spectrum.
            Then, each map cell is \emph{scored} according to its agreement with a reconstructed structural map, brighter parts of the map have a higher structure score.
            The score can be automatically thresholded to \emph{extract} the structural parts of the map (green in the bottom images),
            after which clutter can be \emph{removed}.
            (The symbols are discussed in \cref{sec:dft-for-structure-scoring}.)
        }\label{fig:poster_baby}
    \end{figure}

    Autonomous mobile robots are increasingly being deployed in real-world industrial settings (\ie{} warehouses, factory and hospital automation, stock-taking in retail).
    However, the share of robot deployments is vanishingly small compared to the size of these markets as a whole.\footnote{https://ifr.org/downloads/press2018/Presentation\_WR\_2020.pdf}
    One severe barrier for robot deployment is the cost and engineering labour required for map creation and validation, especially in settings where high precision and repeatability are required.

    A mission-critical capability for a mobile robot is the ability to navigate reliably in its environment;
    this capability requires a useful map.
    Many robots can build and use a map of their environment to navigate and localize via Simultaneous Localization And Mapping (SLAM).
    However, even state-of-the-art SLAM methods~\cite{7747236} can produce maps with misaligned elements and artefacts.
    Moreover, even a nominally correct map on occasions needs to be manually cleaned of clutter before it is used in production.
    Furthermore, good spatial maps are fundamental for semantic understanding,
    \eg{}, for environment partitioning~\cite{Luperto2019}, where the map is split into semantically meaningful parts (rooms).

    In this paper, we present \ac{rose}, a method that automatically extracts the global structure of noisy and cluttered 2D maps, which can be used to improve map quality (by removing clutter) and to detect broken maps.
    While indoor environments can be complex, most human-made environments' main global structure is formed by straight walls.
    Therefore, extracting the building-level structure, by which we mean a configuration of lines, can provide valuable information both about the environment itself and the quality of the generated map.

    The structure is found by extracting key directions from the frequency spectrum of a 2D grid map, which are also used to determine how well each map cell agrees with the structure.
    One application is to remove clutter from maps, and we demonstrate this ability using a set of real-world maps with induced clutter.
    We also show that the structure information can indicate which parts of the map are accurate and which part of the map may require remapping or more careful navigation.
    Finally, we show how a decluttering step can improve wall detection, which is fundamental for, \eg{}, room segmentation~\cite{Luperto2019}.

    \section{Related Work}
    \label{sec:related-work}
    Mobile robots often operate in human-made environments.
    Thus, extracting structure from raw sensor data can benefit map building~\cite{Nguyen2005,Jian2016}.
    However, structure information can be useful also after the map has been constructed.

    Some authors have used the Hough or Radon transform to detect the principal alignment of robot maps, primarily for map matching.
    \Eg,~\cite{saeedi-2014-merging} find peaks in the Hough images of maps to generate rotation candidates for alignment.
    Similarly,~\cite{carpin-2008-merging} finds an alignment of two maps from the Hough \emph{spectra} of the maps.
    However, as we demonstrate in \cref{sec:results}, the Hough transform is not generally suitable for detecting structure in imperfect maps.~\cite{shahbandi-2014-decomposition} use a histogram of oriented gradients (HOG) in combination with radiograms~\cite{bigun-1996-radiograms} and argue that this method can extract dominant elements more robustly than the Radon transform when structural features are corrupted by discontinuities or noise.

    A limitation of the abovementioned methods for structure extraction is that they focus on detecting the most prominent line orientations and require that the number of orientations is known in advance.
    We aim to detect line segments and alignments and facilitate the \emph{reconstruction} of the underlying map structure, given an imperfect input map acquired by a robot.

    The method of~\cite{shahbandi-2014-decomposition} has later been used for rigid and nonrigid multimodal map merging~\cite{shahbandi_2018_alignment,shahbandi-2018-nonlinear}.
    Although the intended application is different, their radiogram-based decomposition, pruned using a distance image of the map's free space~\cite{shahbandi_2018_alignment}, can be said to be a form of structure extraction.
    However, it can only work for clutter-free maps.
    All the above methods are designed to work with clutter-free maps, as opposed to \ac{rose}.
    A critical application of our structure extraction method is to remove clutter from maps, which is
    an essential prerequisite for many other techniques for map segmentation, alignment, etc.

    We also argue that our structure extraction method will be beneficial for reference-free map quality assessment.
    Algorithmically assessing whether a map is correct (and which parts are not) is essential when deploying a robot  (not least for industrial service robots).
    Assessing a map's correctness is a different problem than the evaluation of SLAM methods using standardised benchmarks.
    One example of a framework for assessing map quality is the one of~\cite{Schwertfeger2016}, which produces several scores (coverage, relative accuracy, etc.) of a robot map when compared with a reference map.
    The key idea is to build graphs from the robot map and the reference and compare the two.
    However, because it hinges on comparing the map to a reference, it is of little use when deploying a robot in a new environment.~\cite{Chandran-Ramesh2008} proposed to train a conditional random field (CRF) to assess ``plausible'' and ``suspicious'' configurations of wall segments in 2D maps and demonstrated that this method could be used to label mapping errors without relying on a reference map.
    In comparison, our method is \emph{learning-free} and instead exploits the fact that buildings typically are structured with walls aligned along a limited set of directions.
    However, we are not limited to purely rectangular environments but can also detect structure in cases with multiple directions.

    There is also a line of work that, instead of assessing the (global) structure of the {map}, works on the sensor data and assesses the ``crispness'' of the accumulated points.
    Some authors have used a voxelised representation~\cite{douillard-2012-scan,das-2012-3d} and others have measured the mean point entropy~\cite{droeschel-2018-efficient}.
    These measures are useful for comparing the results of different map algorithms of the same environment but would be challenging to use as a general map score, as the entropy or crispness of a ``good'' map is mainly environment-dependent.

    \section{Structure Scoring with ROSE}
    \label{sec:dft-for-structure-scoring}

    \ac{rose} exploits the fact that indoor environments usually contain walls and straight-line elements along a limited set of orientations.
    Therefore 2D metric maps of such environments often have a set of {\em dominant directions} (see \cref{fig:poster_baby}), that are distinguishable in the frequency spectrum of the 2D binary map.
    We extract these dominant directions as lines from the frequency spectrum and use them to reconstruct the environment's structural elements.

    \subsection{Structure Detection}
    \label{subsec:dominant-direction-identification}
    The first step of \ac{rose} is to compute the 2D \ac{2dft} $\fmap$ of an input map $\map$.
    To extract the dominant directions, we ``unfold'' the amplitude ($\amp=|\fmap|$) of the \ac{2dft} about the center of the plot.
    The unfolding function $c$ estimates the amplitude ($\ampp$) values for an equally spaced grid of orientations ($\phi$) and distances ($\rho$) from the center of the frequency spectrum, based on the input array of the \ac{2dft} amplitudes.

    Once the spectrum is unfolded, we compute the cumulative amplitude $\Camp$ as sums of amplitudes $\ampp$ along each direction $\phi$:$\Camp(\phi)=\sum_{\rho=0}^{N}\ampp(\phi,\rho)$.
    The cumulative amplitude has peaks where the frequencies correspond to lines in the image, as shown in \cref{fig:poster_baby} (second panel from the top).
    The peaks are selected based on their prominence $\mathrm{Pro}(\Camp(\phi))$, which measures the height of a peak relative to the lowest line encircling it but containing no higher peaks within it.\footnote{As implemented in SciPy v1.4.1 \texttt{peak\_prominences} function.}
    A direction $\phi$ is labelled as a peak if its prominence is greater than a predefined threshold $t$.
    We set  $t$  to 50\% of the relative peak height, as this tends to allow keeping the dominant directions in the map while discarding noise in a wide range of real-world maps evaluated in  \cref{sec:results}.

    The set $\pind$ of peak points selected in this way constitutes the dominant directions of the original map $\map$.
    This set can then be used to assess the structure of the map both at the global level and also for individual grid cells.

    \subsection{Global Structure Scoring}
    \label{subsec:global-strucutre-scoring}

    The number and the prominence of the extracted peaks indicate the general level of structure in the assessed map.
    If $|\pind|=0$ peaks are extracted, the map does not contain any dominant directions and is not suitable for further processing.

    To assess the overall level structure of the map (as distorted maps may also contain peaks fulfilling the prominence requirement), we propose to measure how much the cumulative amplitudes of the peak directions $\Camp(\pind)$ are elevated over the cumulative amplitudes $\Camp(\phi)$ of all the directions.
    Considering that the cumulative amplitude depends on the size of the map and amount of occupied cells in it, we apply min-max feature scaling to make the score comparable between maps of different sizes.
    \begin{equation}
        \scampind{\phi} = \frac{\Camp(\phi) - \min(\Camp)}{\max(\Camp)-\min(\Camp)}\label{eq:scaled_scores}
    \end{equation}
    The level of structure is computed as the ratio between the scaled average cumulative amplitude and average peak value,
    \newcommand{\globalscore}{\mathcal{W}}
    \newcommand{\avgsignal}{\overline{\mathcal{A}^{C}_{p}(\phi)'}}
    \newcommand{\avgpeak}{\overline{\mathcal{A}^{C}_{p}(\pind)'}}
    \begin{equation}
        \globalscore = \avgsignal / \avgpeak
        ,
        \label{eq:env_score}
    \end{equation}
    where maps with clear structure have $\globalscore$ close to 0, and maps without structure have $\globalscore$ close to 1.
    As shown in \cref{subsec:scroring-certiffication}, this value can be used as an uncertainty measure of structure detection.

    \subsection{Local Structure Scoring}
    \label{subsec:local-strucutre-scoring}
    After the global structure detection described above,
    the next step is to identify to what extent the occupied map cells belong to the dominant directions.

    For this purpose, we divide the frequency spectrum into two parts, a structured and a non-structured part.
    The structure part ($S$) contains the frequency components along the
    peak directions in $\pind$:
    \begin{equation}
        \label{eq:S}
        S=\{(u,v)_s|(u,v)=c^{-1}(\phi,\rho),\phi\in\!\Phi_p,\rho\in(0,\rho_{\mathrm{max}})\}%
        .\!
    \end{equation}

    To obtain the cells in the frequency spectrum $\fmap$ that correspond to the structure ($S$), we apply the folding function ($c^{-1}$).
    The folding function finds all the cells in the frequency spectrum that share orientations ($\Phi_p$) with the peaks.
    The remaining part of the frequency spectrum  $N=S^{C}$ is then labelled as
    non-structure.

    $S$ is then used to reconstruct the structured elements of the map using the \ac{idft}.
    This constitutes a \emph{nominal reference map} $\nmap$; \ie, a representation of what we expect a ground-truth map to look like, in lieu of an actual reference map.

    \begin{equation}
        \nmap (m,n)=\frac{1}{XY}\sum^{X-1}_{u=0}\sum^{Y-1}_{v=0}\fmap (u,v)e^{j2\pi(um/X+vn/Y)}, (m,n)\in S \label{eq:pix_score}
    \end{equation}

    The pixel score computed in \cref{eq:pix_score} can be further applied to label pixels as part of either structure or clutter.
    The split can be executed through simple thresholding.
    To automatically estimate the threshold value, we propose to use a \ac{gmm}.
    To find the threshold, we first run \ac{em} over the list of pixel scores.
    In this way we obtain two normal distributions: $\mathcal{N}_\mathrm{structure}$ and $\mathcal{N}_\mathrm{clutter}$ (see \cref{fig:struct_score}).
    The threshold is defined as the pixel score $s \in \nmap$ for which the two Gaussians intersect: $\tau\mathcal{N}_{\mathrm{structure}}(s)=(1-\tau)\mathcal{N}_{\mathrm{clutter}}(s)$.

    Finally, the decluttered map is constructed as
    \newcommand{\tmap}{\mu_T}
    \begin{equation}
        \label{eq:tmap}
        \tmap(x,y)=\begin{cases}
                       1 ~&\text{ if }~  \nmap(x,y)> s~, \\
                       0 ~&\text{ otherwise }.
        \end{cases}
    \end{equation}

    \section{Experimental Evaluations}
    \label{sec:results}

    In this section, we demonstrate the performance of \ac{rose} for decluttering.
    First, we present a use-case of using \ac{rose}  with a recent map segmentation algorithm~\cite{Luperto2019}, comparing against a probabilistic Hough transform as a baseline method for extracting structure from cluttered maps (\cref{subsec:environment-segmentation}).
    We also present a sensitivity analysis for a large range of environments and clutter configurations (\ref{subsec:simple-strucutre-extraction}), and a validation of using $\globalscore$ as an uncertainty measure for \ac{rose}'s output (\ref{subsec:general-strucutre-scoring}).

    \subsection{Decluttering for Wall Detection}
    \label{subsec:environment-segmentation}
    \begin{figure}[H]
        \centering
        \subfloat[Input map]{\includegraphics[width=0.3\columnwidth]{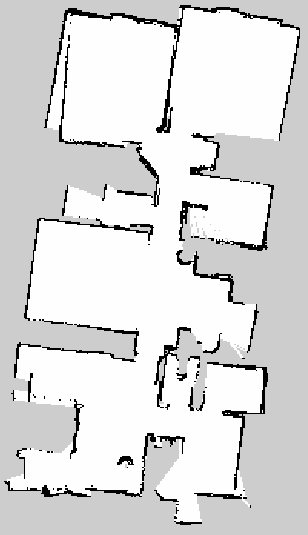}}
        \hfill
        \subfloat[Filtered map]{\includegraphics[width=0.3\columnwidth]{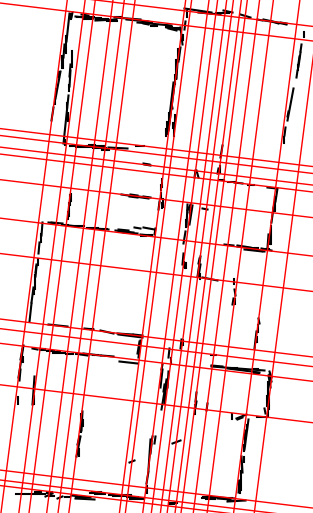}}
        \hfill
        \subfloat[Non-filtered map]{\includegraphics[width=0.3\columnwidth]{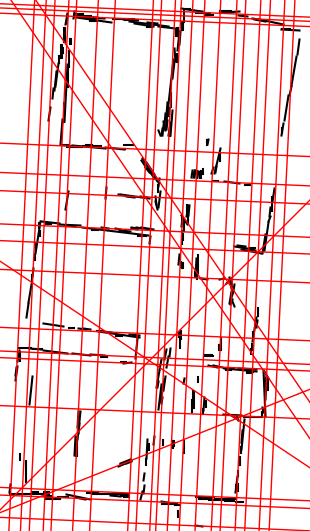}}
        \caption{Wall line extraction performed with and without applying the \ac{rose} decluttering filter in an apartment map (\emph{Aldine}).
        Note that without applying the \ac{rose} decluttering filter, the wall extraction algorithm overestimates the number of walls.}\label{fig:wall_extraction}
    \end{figure}

    In this section, we demonstrate the ability of \ac{rose} to detect building-level structure from cluttered maps by comparing its results with those obtained with Hough transform using as a reference the work of~\cite{Luperto2019}.
    Identifying the underlying structure in indoor environments is usually performed by estimating wall directions using structural features as collinear line segments.
    Typically, the detection of line segments from images as maps is performed using Hough transform, as in~\cite{capobianco2016automatic}.
   However, the Hough transform is not robust when faced with cluttered and noisy maps.
    Depending on the parameter selection, it tends to miss identifying even straight walls when they are partly occluded by furniture or other clutter, or introduce a large number of small line segments from the clutter elements, as shown and discussed in~\cite{armeni20163d}.

    The method of~\cite{Luperto2019} identifies the building-level structure in 2D maps by detecting the \emph{lowest number} of lines that are needed to describe the direction of all walls.
    Wall directions are detected as \emph{wall lines} (in red in \cref{fig:wall_extraction}), obtained after filtering and clustering collinear line segments.
    As line segments are detected from the map $\map$ with the Hough transform, the method suffers from the limitations mentioned above when applied to cluttered maps.
    This is significantly reduced with \ac{rose} by modifying the method of~\cite{Luperto2019} in two ways.

    At first, we use as input the thresholded map $\tmap$, as obtained in \cref{subsec:dominant-direction-identification}, instead of the input map $\map$.
    This allows considering only structural map-features, reducing the initial number of line segments filtering wrongly-identified line segments from clutter.
    Secondly, we align wall lines to dominant directions $\Phi$.
    This reduces the number of lines and resolves alignment issues (e.g. by aligning two near-parallel wall lines in the map).

    Our experimental evaluation is performed by using cluttered real-world maps from small-scale (apartments) to large-scale (offices) indoor environments, from two public datasets (Radish~\cite{radish} and Gibson~\cite{gibson}) and maps of real-world apartments~\cite{ECMR19}.
    Note that structure identification in small-scale apartments is particularly challenging due to furniture occluding walls, as shown in~\cite{armeni20163d}.
    To quantitatively assess the impact of ROSE map filtering for this use-case, we measure the minimal Euclidean transformation between the detected wall lines and manually marked ground-truth wall lines, with and without first applying \ac{rose}.
        This metric allows to measure the offset error and missing or redundant wall detections.

    The Euclidean transformation is computed as $V_{ref}=R(\theta)*V_{\mathrm{test}}+T$, where $V_{\mathrm{test}}$ is a line segment in the test map, $V_{\mathrm{ref}}$ is a line segment in the reference map, $\theta$ and $T$ are the rotation angle and translation vector respectively necessary to align the line segments.
    We compute the cost of transformation as the length of an arc where the length of the translation vector is the radius, and the rotation angle is the arc length $A=\theta|T|/2$.
    The map's score is the average cost for all of the lines detected in the test map.

\begin{figure}[H]
        \begin{tikzpicture}
            \footnotesize
            \begin{axis}
                [xtick = {0,1,2,3,4,5,6,7,8,9,10,11,12,13,14,15}, xticklabels = {Moonachie, Allensville, American, Apartment Y, Aldine, Apartment Q, Freiburg, Almena, Daetsville, Almota, Apartment Z, Apartment X, Alstown, Pomaria, Map15, Map33
                },scatter/classes={g={mark=halfcircle*,draw=green},r={mark=halfcircle*,draw=red}},ymode=log,xticklabel style={rotate=85},
                height=6cm,width=\linewidth,
                ]
                \addplot[scatter,only marks,scatter src=explicit symbolic, mark options={scale=1}]%
                table[meta=label]{
                    x y label
                    0 5.349 r
                    0 0.749 g
                    1 23.986 r
                    1 15.252 g
                    2 7.979 r
                    2 0.029 g
                    3 0.442 r
                    3 0.040 g
                    4 7.086 r
                    4 0.015 g
                    5 1.666 r
                    5 0.020 g
                    6 9.300 r
                    6 0.026 g
                    7 18.481 r
                    7 0.035 g
                    8 0.786 r
                    8 0.099 g
                    9 10.757 r
                    9 0.032 g
                    10 0.727 r
                    10 0.039 g
                    11 2.339 r
                    11 0.046 g
                    12 8.248 r
                    12 0.044 g
                    13 8.399 r
                    13 0.002 g
                    14 6.751 r
                    14 0.010 g
                    15 5.953 r
                    15 0.006 g
                };
            \end{axis}
        \end{tikzpicture}
        \caption{Comparison of the average
        wall-line extraction error for maps with (green) and without (red) filtering with \ac{rose} in cluttered maps. The maps belong to three different data sets: Apartments "X", "Y", "Z" and "Q" are real-world apartments~\cite{ECMR19}; Aldine, Allensville, Almena, Almota, Alstown, American, Daetsvillle, Moonachie, Pomaria come from a dataset containing realistic simulation~\cite{gibson}; and mapped using~\cite{gmapping} for SLAM and Map33, Map15, Freiburg are maps from the Radish dataset~\cite{radish}.
        }\label{fig:wall_ext_plot}
    \end{figure}
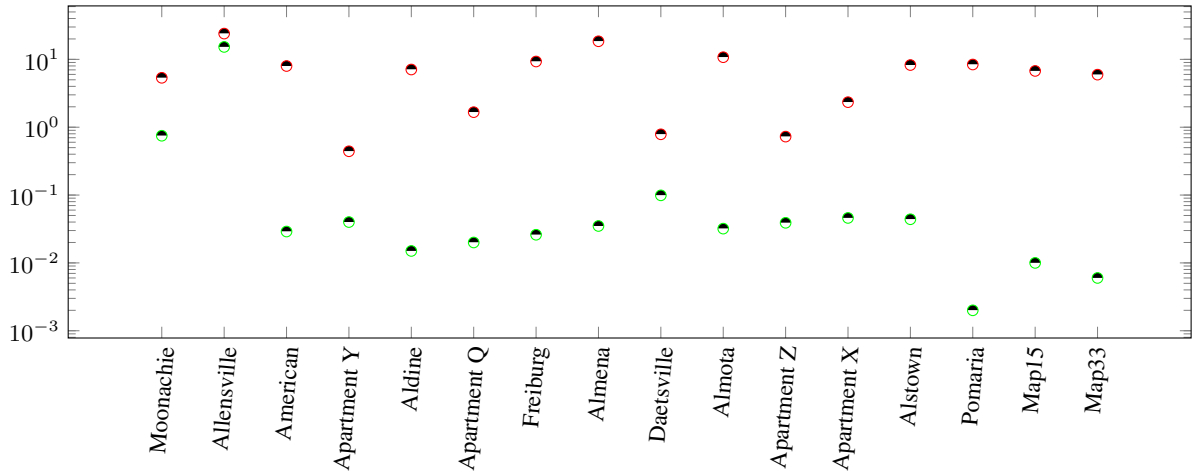

    In \cref{fig:wall_ext_plot} and \cref{fig:wall_extraction}, we can compare the results with and without filtering with ROSE.
    (The plots for remaining environments are shown in the accompanying video.)
    These results show that ROSE significantly improves the robustness of wall detection in cluttered maps by having significantly lower reconstruction error if compared with the Hough transform-based baseline.
    In 15 of the 16 cases, the error is 1-2 orders of magnitude lower.
    In 1 of the 16 cases (the "Allensville" map), the difference is smaller.
    This is a more challenging map, where the error is substantially larger both with and without filtering.

    \subsection{Clutter Sensitivity Analysis}\label{subsec:simple-strucutre-extraction}

    \begin{figure}[H]
        \centering
        \includegraphics[width=\linewidth]{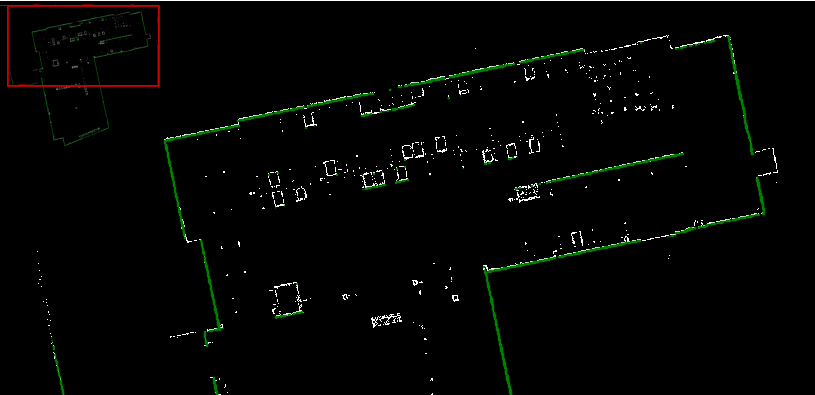}
        \caption{Structure extraction from a real-world robot map.
        \ac{rose} extracts the three dominant directions of the walls in this environment.
        It marks as clutter pixels which, even though they are aligned with the dominant directions, are not prominent enough to be treated as a part of the structure.}
        \label{fig:orkla_tr}
    \end{figure}

    \begin{figure*}[H]
        \includegraphics[width=\textwidth]{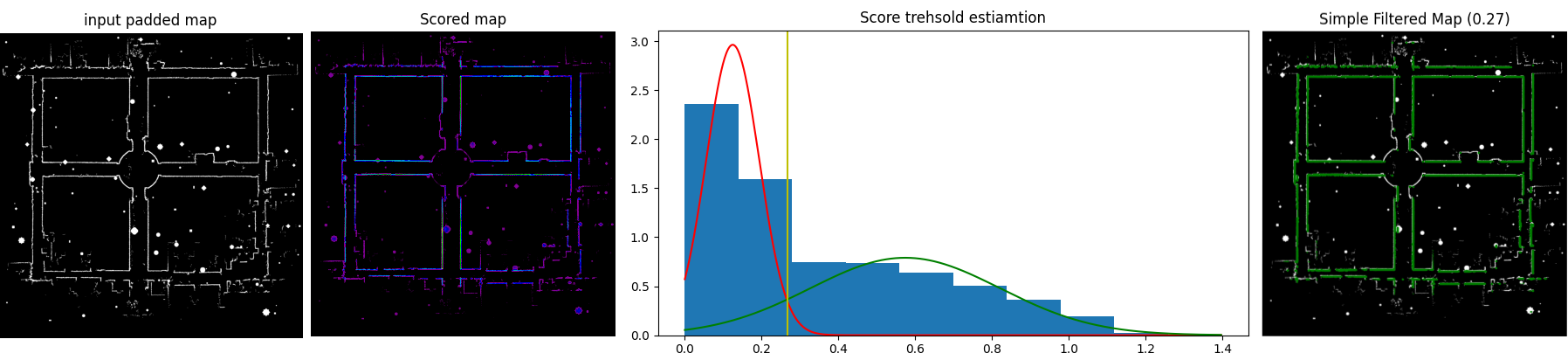}
        \caption{Example of structure labeling based on the structure score.
        Starting from the left, the first figure is an input map.
        The second figure shows the distribution of the pixel quality scores (the darker purple color correspond to lower scores, brighter green and blue correspond to higher scores).
        Then we can see the threshold estimated as the intersection between two Gaussians (The red one is $\mathcal{N}_{\mathrm{clutter}}$ and the green is $\mathcal{N}_{\mathrm{structure}}$).
        Finally, we can see the pixels labelled as "correct" highlighted in green in the last image.}
        \label{fig:struct_score}

    \end{figure*}

\begin{figure}[H]
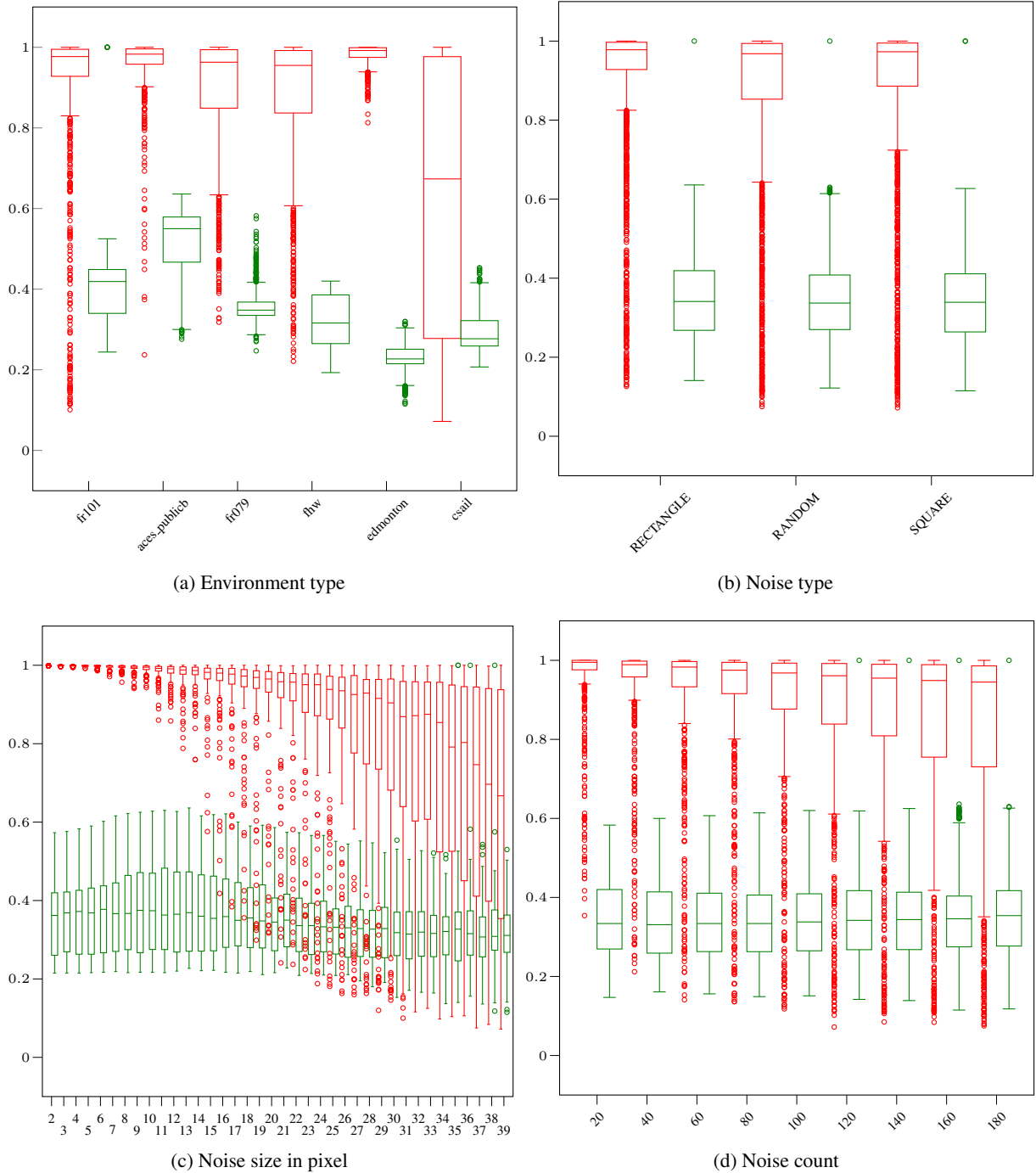

        \subfloat[Environment type]{\includegraphics[width=0.49\columnwidth]{Environment_box.tikz}\label{fig:p_r_env}}
        \subfloat[Noise type]{\includegraphics[width=0.49\columnwidth]{Obstacle_Type_box.tikz}\label{fig:p_r_nt}}\\
        \subfloat[Noise size in pixel]{\includegraphics[width=0.49\columnwidth]{Obstacle_Size_box.tikz}\label{fig:p_r_ns}}
        \subfloat[Noise count]{\includegraphics[width=0.49\columnwidth]{Obstacle_count_box.tikz}\label{fig:p_r_nc}}
        \caption{Precision and recall with respect to different noise parameters (Figs. b, c, d aggregate results from all the environments.). Precision is shown in red, and recall in green.}
        \label{fig:p_r}
    \end{figure}

    In this section we evaluate the capabilities of \ac{rose} to declutter noisy maps providing a quantitative analysis of \ac{rose} structure scoring performance w.r.t. different amounts and types of clutter (see \cref{fig:p_r}).

    We have selected six maps from the Radish dataset~\cite{radish} representing different complexity levels and added additional artificial noise.
    The map names are shown in \Cref{fig:p_r_env}.
    Due to space limitations, the maps are shown in the video attachment to the paper.
    Following the idea presented by~\cite{bormann-2016-segmentation},
    we have added artificial clutter of different size, shape and amount.
    We have added clutter shaped as squares, rectangles, and random shapes (circles, diamonds, stars).
    We also have graded the amount (from 20 to 180 obstacles per map) and size (from 2 to 39 pixels) of the added obstacles.

    We process the cluttered maps with \ac{rose} structure scoring, and select ``structure'' vs ``clutter'' using a threshold found with the GMM method Eq.~\eqref{eq:tmap}.
    As ground truth, we assume that all occupied pixels in the original maps from the dataset are related to structure and that only the artificial objects are clutter.
    However, please note that this ground-truth labelling is somewhat skewed.
    As these are real-world maps, they also contain some amount of clutter and errors.
    A perfect clutter removal may also remove parts that we have labelled as structure, manifested as a decreased recall rate.
    With that said, 100\% recall means that no structure has been labelled as clutter, and 100\% precision means that no clutter has been labelled as structure.

    \Cref{fig:p_r_env} shows that the median precision is constantly $>$95\% for all environments
    except for \emph{csail}, which is a peculiar building with few straight walls.
    The recall is typically around 30\%, which may seem surprisingly low.
    However, this comes from the problem with finding the correct ground-truth labelling described above.
    An example is shown in \cref{fig:struct_score} (the rightmost image).
    This is the \emph{aces\_publiclib} map, which is relatively clean.
    However, there are several cases of small ``clutter'' segments in partially observed regions near the edges of the map, which are diligently filtered by \ac{rose} even though they are assumed to be ``structure'' when computing the recall.
    (Note that each box plot shows the outcome of \emph{all combinations} of clutter in one environment: 3 $\times$ 39 $\times$ 6 $=$ 1053 data points.)

    \Cref{fig:p_r_nt} shows that the clutter's shape has no substantial impact on the quality of the labelling.
    Both precision and recall are practically the same for all shapes.

    Finally, in \cref{fig:p_r_ns,fig:p_r_nc}, we can see that the size and the number of obstacles affect the quality of the labelling the most.
    While the recall is mostly constant, the median precision starts to decrease noticeably when more than 80 obstacles, or obstacles larger than 15 pixels in size, are added to the maps.
    In other words: as more and larger pieces of clutter are added, at some point, they start to appear more like structure.

    \subsection{Confidence Measure for Structure Scoring}
    \label{subsec:general-strucutre-scoring}

    \begin{figure}[H]
        \raisebox{3mm}{\includegraphics[width=0.19\linewidth]{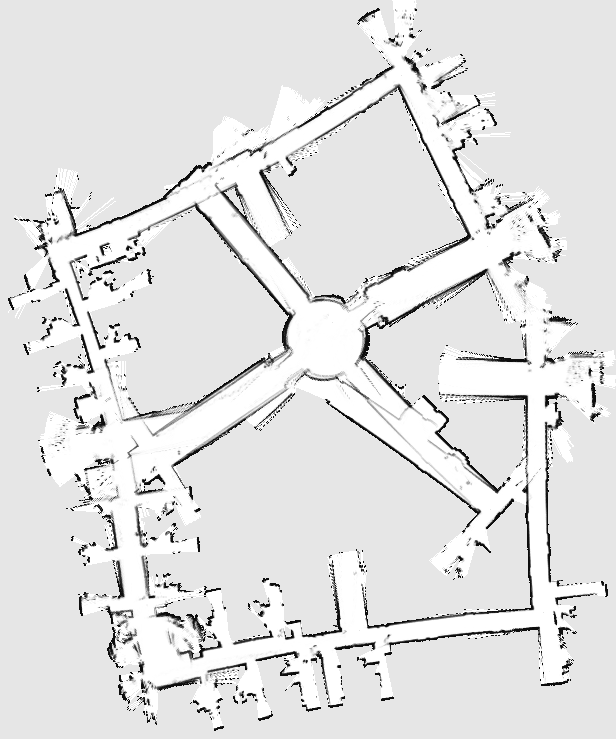}}
        \includegraphics[width=0.29\linewidth]{aces_publicb_bad.tikz}
        \raisebox{4mm}{\includegraphics[width=0.19\linewidth]{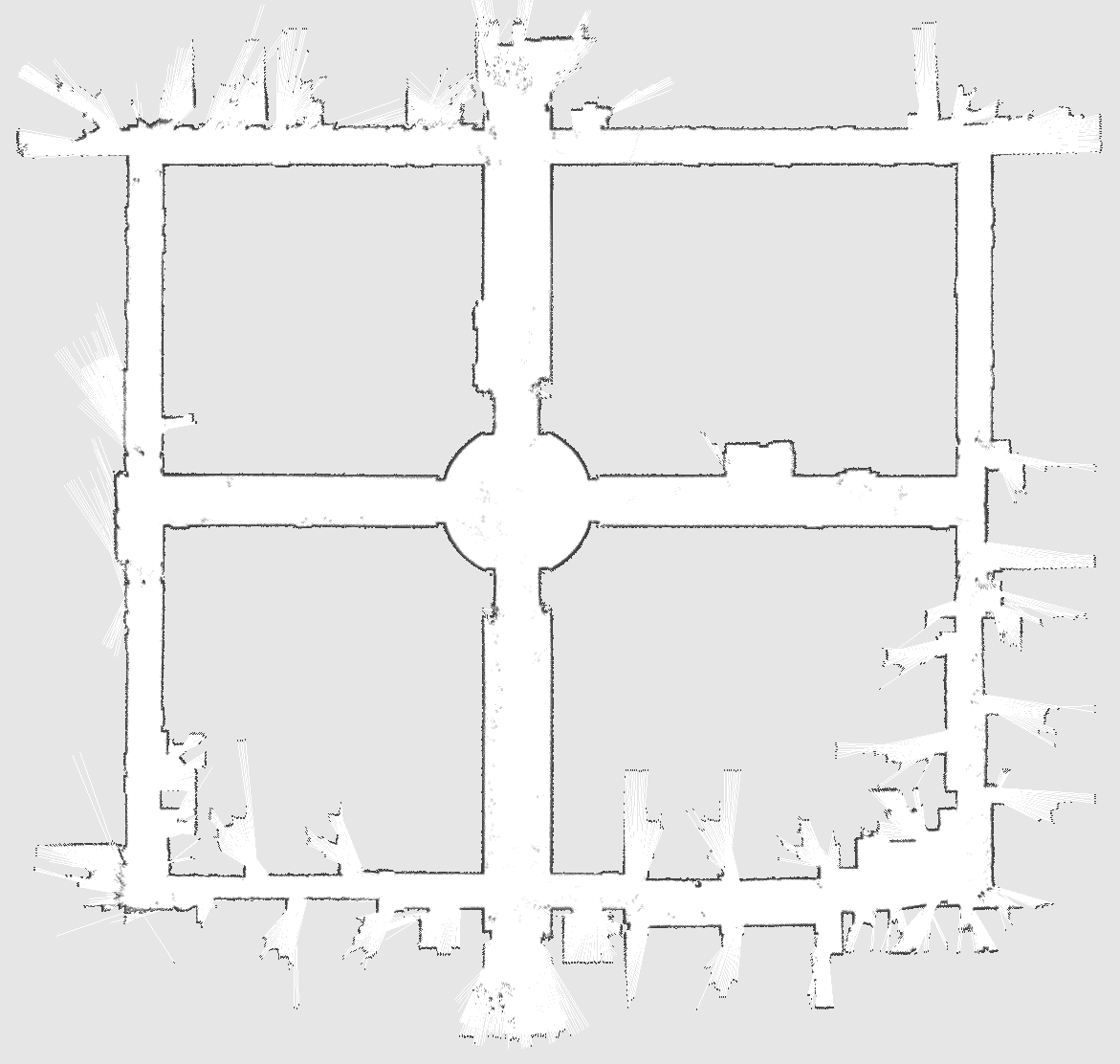}}
        \includegraphics[width=0.29\linewidth]{aces_publicb_good.tikz}
        \caption{Comparison of the $\globalscore$ measure between the correct and the distorted map; \ie, the ratio between the average signal value $\avgsignal$ (orange line) and the average peak value $\avgpeak$ (green line).
        For the map on the right $\globalscore= 0.07$ and for the map on the left $\globalscore= 0.47$.}
        \label{fig:structure_scores}
    \end{figure}

    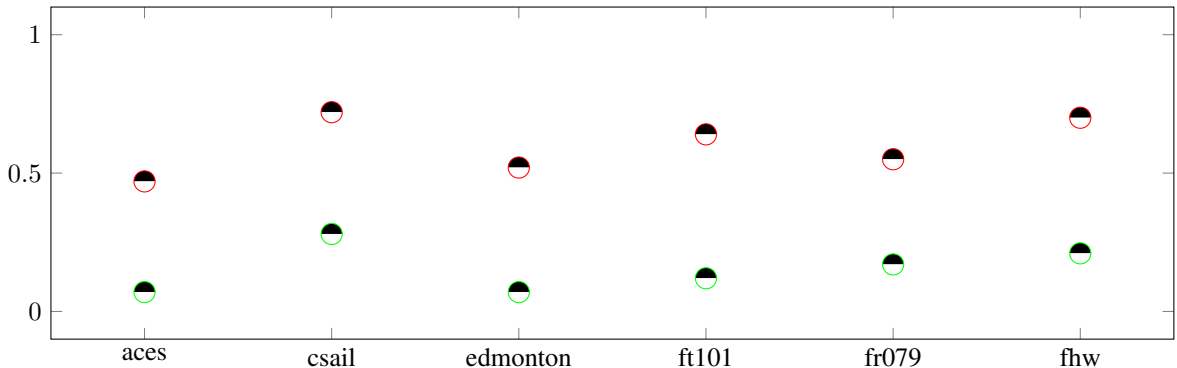
\begin{figure}[H]
        \centering
        \begin{tikzpicture}
            \begin{axis}
                [xtick = {0,1,2,3,4,5}, xticklabels = {aces,csail,edmonton,ft101,fr079,fhw},scatter/classes={g={mark=halfcircle*,draw=green},r={mark=halfcircle*,draw=red}},ymin=-0.1,ymax=1.1, height=6cm,width=\linewidth]
                \addplot[scatter,only marks,scatter src=explicit symbolic, mark options={scale=2}]%
                table[meta=label]{
                    x y label
                    0 0.07 g
                    1 0.28 g
                    2 0.07 g
                    3 0.12 g
                    4 0.17 g
                    5 0.21 g
                    0 0.47 r
                    1 0.72 r
                    2 0.52 r
                    3 0.64 r
                    4 0.55 r
                    5 0.70 r
                };
            \end{axis}
        \end{tikzpicture}
        \caption{Difference between the structure score $W$ for maps depending on their quality. The red markers correspond to broken maps (built using only wheel odometry) and the green ones correspond to correct maps (with SLAM).}\label{fig:gen_struc_score}
    \end{figure}

    In \cref{subsec:global-strucutre-scoring} we proposed a measure $\globalscore$  of the global amount of structure present in a map.
    \Cref{fig:gen_struc_score,fig:structure_scores} demonstrate a use-case for applying this measure to detect whether or not a map is consistent.

    \Cref{fig:structure_scores} illustrates the correct and a distorted map, and the corresponding amplitude plots used to compute $\globalscore$.
    The correct maps are results of graph-SLAM optimisation, and the distorted maps are built using only the odometry.
    For the distorted map, the $\globalscore$ measure is substantially higher than in the correct map.

    \balance
    \Cref{fig:gen_struc_score} compares $\globalscore$ for correct maps (green markers) and distorted maps (red markers) from RADISH dataset~\cite{radish}.
    We can observe that the $\globalscore$ for correct maps is substantially lower than for the maps built only with odometry.
    That shows that $\globalscore$ can indicate the amount of structure in the environment: lower in distorted maps.

    \subsection{Scoring certification}
    \label{subsec:scroring-certiffication}
    \begin{figure}[H]
        \includegraphics[width=\columnwidth]{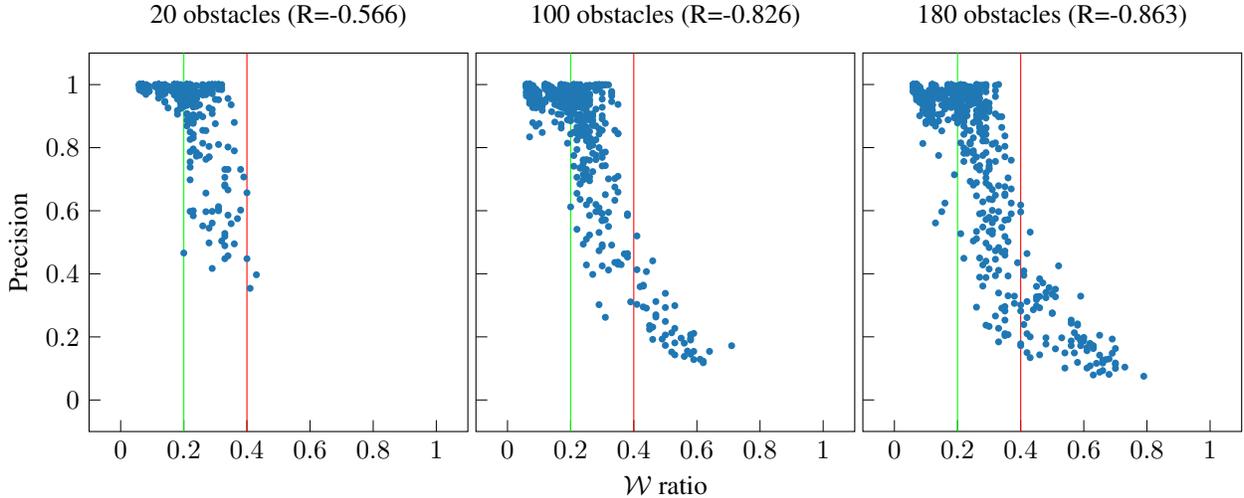}
        \caption{%
            Use of $\globalscore$ as an uncertainty measure for clutter labelling for various numbers of obstacles (noise count).
            Each plot shows the labelling precision vs the ratio $\globalscore$ for all configurations in \Cref{fig:p_r} for one value of noise count; \ie, 702 data points.
            $\globalscore$ is strongly correlated to the precision.
        }
        \label{fig:occ_corr}
    \end{figure}

    In addition to indicating distorted maps (as in \cref{fig:structure_scores}), $\globalscore$ can also be used as a confidence measure when using \ac{rose} for decluttering (via local structure scoring).
    In this experiment, we have used the same maps as in \cref{subsec:simple-strucutre-extraction}.
    \cref{fig:occ_corr} plots $\globalscore$  (horizontal axis) against the precision of clutter labelling (vertical axis).
    These results indicate a strong correlation ($R<-0.7$) between $\globalscore$ and the performance of decluttering.

    In consequence, this measure can be used for self-certification of the labelling.
    As a rule of thumb, $\globalscore<0.2$ means that the decluttering result can be trusted  (high precision) and $\globalscore>0.4$ means that it has almost certainly failed.

    \section{Conclusions \& Future Work}
    \label{sec:conclusions}
    In this paper, we have presented the \ac{rose} method for structure extraction and clutter removal.
    Structure scoring is a method for detecting building-level structure in a map and assessing how much each part of the map agrees with the environment's structure.
    Structure scoring exploits the observation that sets of straight lines in the map have a characteristic representation in the frequency spectrum.

    The work presented in this paper offers a novel approach to the problem of structure extraction.
    In contrast to existing methods, it does not rely on any prior knowledge
    (\eg, the number of dominant directions in the environment)
    and is highly robust to clutter and mapping errors.
    Instead, it uses the general characteristics of the environment.
    \ac{rose} can be used in different applications.
    The fact that \ac{rose} does not rely on learning and requires very little prior knowledge makes the method flexible and applicable for new scenarios and environments without re-training.

    In the presented experiments, we have shown that the proposed method can improve extracting building-level structural elements.
    Using our structure extraction to remove clutter as illustrated in \cref{fig:poster_baby,fig:orkla_tr} also enables better structural features and room identification, which could be later used for room segmentation.
    The map in \cref{fig:poster_baby} is from a recent data set and survey paper on room segmentation~\cite{bormann-2016-segmentation}, where all state-of-the-art methods performed poorly on this and other cluttered maps.

    A limitation of \ac{rose} is that it strongly relies on the idea of structure as an organization of straight walls.
    As such, it is not currently applicable for environments with many curved walls.
    Furthermore, \ac{rose} focuses on the detection of
    structured parts of the environment, which means if the noise is systematical - \eg, offsetting the position of the walls with a fixed value - the ``fake structure'' will be scored equally high as the true one.

    However, it is important to mention that \ac{rose} can self-assess the scoring process' quality.
    The strong negative correlation between the uncertainty measure $\globalscore$ and the precision of labelling shows that it is possible to indicate if such a ``fake structures'' are also labelled as part of the map.

    In this paper, we have covered the critical applications of \ac{rose} for structure extraction in 2D maps.
    Future work can further develop along two key directions.
    The first direction is to extend the capabilities of the method.
    That includes application of \ac{rose} to 3D maps and further development of filtering of the frequency spectrum.

    The second direction is to further investigate the application areas of \ac{rose}.
    For example, how structure scoring can be incorporated into SLAM approaches for improving scan registration and graph optimization.
    As well as, how \ac{rose} can be used to improve room segmentation, semantic understanding and post-processing of cluttered maps.

    \bibliography{ROSE}

\end{document}